\documentclass[a4paper,11pt]{article}

\usepackage[T1]{fontenc}

\usepackage{graphicx}
\usepackage{epsfig,amsmath,amssymb,url}

\renewcommand{\epsilon}{\varepsilon}

\begin{document}

\title{From knowledge-based to data-driven modeling of fuzzy rule-based systems: A critical reflection}

\author{Eyke H\"ullermeier\\
              Department of Computer Science\\
              Paderborn University\\
              eyke@ubp.de    
}

\date{October 2017}


\maketitle

\begin{abstract}
This paper briefly elaborates on a development in (applied) fuzzy logic that has taken place in the last couple of decades, namely, the complementation or even replacement of the traditional \emph{knowledge-based} approach to fuzzy rule-based systems design by a \emph{data-driven} one.  It is argued that the classical rule-based modeling paradigm is actually more amenable to the knowledge-based approach, for which it has originally been conceived, while being less apt to data-driven model design. An important reason that prevents fuzzy (rule-based) systems from being leveraged in large-scale applications is the flat structure of rule bases, along with the local nature of fuzzy rules and their limited ability to express complex dependencies between variables. This motivates alternative approaches to fuzzy systems modeling, in which functional dependencies can be represented more flexibly and more compactly in terms of hierarchical structures. 
\end{abstract}

\section{Introduction}

Since their inception about 50 years ago, marked by Lotfi Zadeh's seminal paper \cite{zade_fs65}, and rapid emergence in the following decades, fuzzy sets and fuzzy logic have found their way into numerous fields of application, such as engineering and control, operations research and optimization, databases and information retrieval, data analysis and statistics, just to name a few.

While different tools from fuzzy logic and fuzzy set theory (FST) have been employed in all these fields, it is fair to say that \emph{fuzzy rule models} or \emph{fuzzy rule-based systems} (FRBS) have received special attention. Indeed, rule-based models have always been a cornerstone of fuzzy systems and a central aspect of research in fuzzy logic---the term ``fuzzy system'' is mostly even used as a synonym for fuzzy rule-based system. To a large extent, the popularity of rule-based models can be attributed to their potential comprehensibility, a distinguishing feature and key advantage in comparison to ``black-box'' models such as neural networks.

Fuzzy systems provide an interface between humans and machines: Mathematical concepts such as fuzzy sets and generalized logical operators allow for an adequate formalization of vague cognitive concepts and linguistic expressions such as ``high temperature''. Moreover, they provide suitable means for reasoning with such concepts in a meaningful way. In principle, machines and computers thus become amenable to human expert knowledge \cite{fzzy_ctrl}. The importance of rule-based methods can be explained by the fact that human experts often find it convenient to describe their knowledge in terms of IF--THEN rules, which typically connect the values of a set of (independent) input variables and those of one or more (dependent) output variables. 

While corresponding aspects of knowledge representation and reasoning have dominated research in fuzzy logic for a long time, problems of {\em automated learning and knowledge acquisition} have more and more come to the fore in recent years \cite{mpub121}. There are several reasons for this development, notably the following. First, caused by the awareness of the well-known ``knowledge acquisition bottleneck'' and the experience that a purely knowledge-based approach to systems design is difficult, intricate, and tedious most of the time, there has been an internal shift within fuzzy systems research from \emph{modeling} to \emph{learning} and \emph{adaptation}, i.e., from the knowledge-based to the data-driven design of fuzzy systems  \cite{babu_fm}. In fact, the latter not only suggests itself in applications where data is readily available, but can sometimes even be essential. In learning on data streams, for example, models are not only constructed once (from a static ``batch'' of data) but need to be updated continuously in an online manner \cite{ange_ei}, which cannot be accomplished by a human expert. 
Second, this trend has been further amplified
by the great interest that the field of knowledge discovery in
databases and its core methodological components, machine learning and data mining,
have attracted in recent years \cite{fayy_fd96}. Learning from data and data analytics have become ubiquitous topics in the era of ``big data''.

This short paper\footnote{\cite{mpub327} is an expanded version of this paper.} starts with a critical consideration of the transition from knowledge-based to data-driven fuzzy modeling. The basic claim we make is that fuzzy (rule-based) systems, which have originally been introduced for the knowledge-based design of relatively small or at best moderately sized models, are not very apt to the data-driven approach to model construction: they are not really appropriate for large-scale modeling and tend to be inferior to mainstream machine learning methodology for model induction and adaptation. In our opinion, the fuzzy rule-based approach is often adopted in a too uncritical way and is not sufficiently questioned in the research community. Indeed, apart from several indisputable advantages, this approach has a number of potential disadvantages. Two reasons for our reservation will be detailed in the paper:
\begin{itemize}
\item First, due to their flat structure, standard fuzzy (rule-based) systems are not scalable and hardly able to capture complex dependencies with many input variables, which are typical of nowadays data-intense applications.
\item Second, fuzzy systems automatically extracted from data easily lose one of their main advantages, namely, their interpretability and cognitive plausibility. 
\end{itemize}

\section{Knowledge-based versus data-driven fuzzy modeling}


\subsection{Knowledge-based modeling}

The knowledge-based approach is at the origin of fuzzy rule-based systems and closely connected to the classical expert systems paradigm: A human expert seeks to formalize her knowledge about the relationship between certain variables of interest (for example, a control function mapping system states to control actions) using IF--THEN rules,  taking advantage of fuzzy sets as a convenient interface between a qualitative, symbolic and a quantitative, numerical level of knowledge representation.  As an illustration, consider a recent application from textile industry, namely, the modeling color yield ($K/S$) in polyester high temperature dyeing as a function of disperse dyes concentration (\texttt{conc}), temperature (\texttt{temp}) and time. The human expert describes this dependency using the following rules \cite{mpub227}:
\begin{itemize}
\item[1.] If \texttt{temp} is low and \texttt{time} is low and \texttt{conc} is low, then K/S is very low.
\item[2.] If \texttt{temp} is medium and \texttt{conc} is high, then K/S is high.
\item[3.] If \texttt{temp} is high and \texttt{conc} is low, then K/S is medium.
\item[4.] If \texttt{temp} is low and \texttt{time} is high and \texttt{conc} is low, then K/S is very low.
\item[5.] If \texttt{temp} is high and \texttt{conc} is high, then K/S is very high.
\item[6.] If \texttt{temp} is medium and \texttt{time} is low and \texttt{conc} is high, then K/S is medium.
\item[7.] If \texttt{temp} is medium and \texttt{time} is high and \texttt{conc} is high, then K/S is high.
\item[8.] If \texttt{temp} is low and \texttt{time} is low and \texttt{conc} is high, then K/S is low.
\end{itemize}
A set of informal rules of that kind is then translated into a mathematical model $M$ by 
\begin{itemize}
\item
assigning a fuzzy subset to each linguistic term (such as ``high temperature''), and thereby a fuzzy partition for each variable,
\item choosing a generalized logical operator for conjunction (``and'') and implication (``then''), 
\item 
specifying an inference procedure such as Mamdani-Assilian \cite{mamd_ae75} or Takagi-Sugeno inference \cite{taka_fi85},
\item 
defining a fuzzification and a defuzzification procedure (mapping, respectively, non-fuzzy to fuzzy and fuzzy to non-fuzzy quantities).
\end{itemize}
A can be seen, a fuzzy model $M$ of that kind is highly ``parameterized''  and involves many degrees of freedom: the structure of the rules, the parameters of fuzzy sets, etc. Eventually, the system as a whole  realizes a mapping 
$$
f_M: \, \mathbb{R}^d \longrightarrow \mathbb{R} \enspace .
$$
In the case of the above example, this function would simply map each input triple $(\texttt{conc}, \texttt{temp}, \texttt{time}) \in \mathbb{R}^3$ to a number $K/S \in \mathbb{R}$.


Specifying a fuzzy system in this way is often a difficult and tedious task. Even if the expert is able to communicate her knowledge in an appropriate form, and willing to accept standard choices for fuzzification and defuzzificiation as well as logical operators and fuzzy inference, the specification of a complete and sufficiently consistent knowledge base, consisting of the rules and fuzzy sets, is quite demanding. In particular, the number of rules may become very large. In our above example, this number is still rather small because the number of input attributes is limited to only three. In general, however, the number of rules quickly grows with the number of attributes---a point that we shall come back to further below. Finally, ``tuning'' the system in case it does not immediately realize the desired input/output relationship is difficult, too, because the different components of the system (fuzzy sets, rules, logical operators, (de-)fuzzificiation procedures) are interacting in a complicated and highly non-linear way.

\subsection{Data-driven modeling}

Given these difficulties, it is hardly surprising that data-driven approaches to fuzzy systems modeling have been considered as an alternative. In fact, this alternative suggests itself in case data about the process to be modeled is available. Imagine, for instance, that a set of $N$ measurements 
$$
(\boldsymbol{x}_n , y_n) = (\texttt{conc}_n, \texttt{temp}_n, \texttt{time}_n, K/S_n) \in \mathbb{R}^3 \times \mathbb{R}
$$ 
is available in our above example. One could then be tempted to ``fit'' a fuzzy model $M$ to this data, that is, to search for an instantiation of the model components such that the induced mapping $f_M$ reproduces the data sufficiently well (i.e., $f_M(\boldsymbol{x}_n) \approx y_n$ for $n=1, \ldots , N$).

Obviously, this process of reverse-engineering a fuzzy system bears a close resemblance to standard statistical regression analysis. Seen from this perspective, a fuzzy system can simply be considered as a (non-parametric) regression function. This view becomes especially apparent for certain types of fuzzy systems, for example, systems with Gaussian fuzzy sets and Takagi-Sugeno inference, which are formally more or less equivalent to common regression techniques such as radial basis function (RBF) networks.

Fuzzy systems are of course not limited to the representation of regression functions but can also be used for classification, that is, for implementing functions with a categorical output. In this case, the consequent of single rules is usually a class assignment, i.e., a singleton fuzzy set.\footnote{More generally, a rule consequent can suggest different classes with different degrees of certainty.} Evaluating a rule base (\`a la Mamdani-Assilian) thus becomes trivial and simply amounts to ``maximum matching'', that is, searching the maximally supporting rule for each class. Thus, much of the appealing interpolation and approximation properties of fuzzy inference gets lost, and fuzziness only means that rules can be activated to a certain degree. There are, however, alternative methods that combine the predictions of several rules into a classification of the query \cite{cord_at98}.

A fuzzy model consists of two types of components, a qualitative one that determines the structure of the model and essentially corresponds to the (linguistic) rules, and a quantitative one that comprises all numerical parameters of fuzzy sets, logical operators, etc. A plethora of strategies have been developed for determining these components in a data-driven way. Especially important in this regard are hybrid methods that combine fuzzy logic with other ``soft computing'' methodologies, notably evolutionary algorithms and neural networks. For example, evolutionary algorithms are often used in order to optimize (``tune'') a fuzzy rule base or for searching the space of potential rule bases in a (more or less) systematic way \cite{cord_ty04}. The combination of fuzzy systems with neural networks leads to so-called {\em neuro-fuzzy} systems \cite{jang_aa93,nauc_fo}. 

\subsection{Combined knowledge-based and data-driven modeling}

In addition to a purely knowledge-based and a purely data-driven approach, there is of course also the possibility to combine these two. For example, a human expert may specify the qualitative part of a fuzzy model by providing a set of linguistic rules, whereas the quantitative part is determined automatically by an optimization method that fits the structure to a given set of data. In other words, the data is used to ``calibrate'' the structure of the model as specified by the expert. In the context of the colour yield application mentioned above, this approach has been used successfully in \cite{mpub273}. 

From a (machine) learning point of view, the part of the model that is pre-defined by the expert serves as a restriction of the underlying model space, i.e., the set of candidate models the learning algorithms can choose from (or, using machine learning terminology, it incorporates an \emph{inductive bias}). A restriction of that kind can be very useful, as it may help the algorithm to learn more quickly and avoid the risk of overfitting the data. On the other side, if not being completely correct, the bias incorporated by the expert may also prevent the algorithm from finding the truly best model. In general, the importance of incorporating expert knowledge increases with the sparsity of the data. If data abounds, it can compensate for a lack of expert knowledge, and purely data-driven approaches are often superior to knowledge-based or hybrid alternatives, at least in terms of accuracy.

\section{Limitations of fuzzy rule systems}

With the ever increasing availability of data, fostered by technological advances in data acquisition, storage and management, the data-driven approach to systems design has become more and more prevalent in the previous years, not only in the field of fuzzy logic but in artificial intelligence (AI) in general---this is why (inductive) machine learning methodology has gained in importance as compared to more classical AI topics of knowledge representation and (deductive) reasoning. 

Given the original motivation of fuzzy modeling as a means for expressing expert knowledge, one may wonder whether this methodology is equally apt to the data-driven paradigm and, not less importantly, to what extent it may usefully complement the arsenal of alternative statistical and machine learning methods, including neural networks, support vector machines, and decision trees, amongst others. This question especially arises since standard machine learning methods are ame\-nable to specific optimization techniques (such as backpropagation in the case of neural networks or quadratic programming in the case of support vector machines), and therefore much more efficient from an algorithmic point of view. Indeed, the development of a machine learning method normally involves algorithmic and computational considerations from the very beginning. In the case of fuzzy systems, these aspects became relevant only later on, and because of their complex structure, the only way to identify fuzzy models is via general-purpose optimization tools  like the ``big hammer'' of evolutionary algorithms.

Since we have been discussing the role of fuzzy logic in machine learning on a quite general level elsewhere \cite{mpub121,mpub223}, this topic will not be deepened any further in this article. Instead, because our focus here is on fuzzy systems in the sense of fuzzy rule models, the remainder of this section is devoted to a brief discussion of two issues directly related to the use of such systems for learning and data-driven model construction.

\subsection{The problem of a flat structure}

Our above color yield example is a very simple, low-dimensional model with three inputs and one output variable. In spite of its practical relevance, most applications will typically involve much more inputs---in modern, data-intense applications, it is not uncommon to deal with hundreds or thousands of variables. But even if the number of variables goes beyond a handful, rule-based systems may become problematic \cite{jin_fm00}. A major reason that prevents such systems from scaling to more complex applications is their flat structure. 

Rules are purely local entities that only cover a small portion of the data space (in the form of more or less axis-parallel rectangles) and that are not able to flexibly exploit specific dependencies or independencies between individual variables. Therefore, a large number of such rules is typically needed in order to describe a global relationship between input and output variables. Regardless of whether grid-based methods (starting from fuzzy sets on the individual dimensions and defining rules as products of these sets) or covering techniques (starting from multi-dimensional rules/clusters in the input space and defining  fuzzy sets as one-di\-men\-si\-onal projections) are used, the number of rules will grow exponentially or at least almost exponentially with the dimensionality of the input space. 

Successful tools for large-scale modeling and learning, such as graphical models and deep neural networks, distinguish themselves by the ability to capture (partial) independencies between variables and/or by conquering complexity through abstraction and hierarchical structuring. This is in contrast to the flat, grid-like structure of standard (fuzzy) rule bases. 

Since a fuzzy system eventually represents a (real-valued) function, it is of course possible to combine fuzzy rule-based modeling with generic hierarchical decomposition techniques. For example, suppose a function $f(\cdot)$ with three input variables can be expressed in the following form: 
$$
f(x,y,z) = g(x,h(y,z)) 
$$
Obviously, both $g(\cdot)$ and $h(\cdot)$ can then be expressed in the form of fuzzy rule bases, giving rise to two fuzzy systems with two-dimensional input space instead of one such system with three inputs. In fact, what is thus obtained is a specific type of \emph{hierarchical fuzzy system}---various other types have been proposed in the literature, see  \cite{torra} for an overview. 

Although hierarchical modeling techniques may to some extent overcome or at least alleviate the problems mentioned above, it seems that hierarchical fuzzy systems have their own drawbacks and have not been widely adopted in practice so far. In fact, designing such systems, whether in a knowledge-based or a data-driven way, is not an easy task. Moreover, there is a risk of further compromising the interpretability of fuzzy systems, which, as will be argued in the next section, is an issue already in the case of standard (flat) fuzzy rule bases. 

Another type of hierarchical fuzzy system, called \emph{fuzzy pattern trees}, has been proposed in \cite{huang:patterntrees,ptreg,mpub214,fast}. In contrast to rule-based systems, pattern trees exhibit an \emph{inherently} hierarchical structure. Moreover, each ``inner node'' of the hierarchy is realized in the form of a simple (and easily interpretable) aggregation function (instead of a set of rules wrapped in a fuzzification and defuzzification procedure).

\subsection{The myth of interpretability}\label{p2}

Interpretability is one of the core arguments often put forward by fuzzy scholars in favor of fuzzy models, regardless of whether these models have been constructed in a knowledge-based or a data-driven way. Unfortunately, this argument is not well supported from a scientific point of view, which is partly due to the difficulty of measuring interpretability and model transparency in an objective manner. Instead, the claim that rules are more understandable than ``formulas'' appears to be a commonplace that is simply taken for granted. Without denying the potential usefulness of fuzzy logic in constructing interpretable models in general, we are convinced that current methods for data-driven construction of fuzzy models produce results that are not at all more interpretable than any other type of model, and often even less. There are several reasons for this scepticism. 

A first problem is connected to the previous discussion and concerns the size of practically relevant models. Even if a single rule might be understandable, a rule-based model with a certain level of accuracy will typically consist of many such rules, which may interact in a non-trivial way and might be hard to digest as a whole. On top of this, fuzzy models often allow for rule weighing and may involve complicated inference schemes for aggregating the outputs of individual rules into an overall prediction. Of course, the problem of complexity is shared by other, non-fuzzy methods, too. For example, decision trees are often praised as being highly interpretable, also in mainstream machine learning. This might indeed be true as long as trees are sufficiently small. In real applications, however, accurate trees are often large, comprising hundred of nodes. Again, interpretability is highly compromised then.

When fuzzy sets are constructed in a data-driven way, it is not at all clear that these sets can be associated with meaningful linguistic labels---let alone labels the semantic interpretation of which will be shared among different users. In fact, one should realize that the fuzzy sets produced are strongly influenced by the data set, which is a random sample, and fuzzy partitions might be strongly influenced by outliers in the data. Moreover, fuzzy sets are in the first place tuned toward good approximation and accurate predictions of the function $f_M$ implemented by the fuzzy model, and not toward meaningful semantics.

Likewise, it is rarely discussed in which way a model is eventually presented to the user. Are the fuzzy sets specified in terms of their membership functions, in which case the user might be overloaded with technical details, or is it just presented in a linguistic form? As mentioned before, the latter presupposes an appropriate assignment of linguistic labels to fuzzy sets, which is difficult to establish.

For these reasons, the (alleged) interpretability of fuzzy systems cannot simply be transferred from know\-ledge-based to the case of data-driven fuzzy modeling. In fact, as an important difference between these approaches, one should notice that, in the data-driven approach, the human is no longer at the origin of the model but changes her role from the ``producer'' to the ``consumer'' of a model. Obviously, a model constructed by the expert herself is very close to what she has in mind. Besides, models of that kind are typically small, comprising only a few input variables. Therefore, one can indeed suppose such models to be understandable. These properties are no longer valid in the case of a data-driven approach, however.

\section{Conclusion}

Fuzzy modeling has originally been conceived as a human/machine interface, namely, for translating informal rules involving linguistic terms with vague semantics into precise mathematical models amenable to computerized information processing. In the spirit of the classical expert systems paradigm, which has dominated research in AI in the 80th of the last century, fuzzy systems have been used quite successfully in many applications. More recently, however, the interest has shifted from the traditional knowledge-based approach to systems design to a data-driven one, that is, the automatic extraction of fuzzy models from data. 

Although this development is a coherent reaction to the increased availability of data and completely in line with the growing importance of machine learning, we suspect that traditional fuzzy systems lose most of their merits when being designed in a purely data-driven manner. Moreover, as a tool for model induction and predictive modeling, fuzzy approaches have a hard time competing with modern machine learning methodology, especially with regard to algorithmic aspects, computational efficiency, and theoretical foundations (guarantees for generalization performance, support of model selection, etc.). The scalability of fuzzy systems is severely hampered by the flat structure of standard rule bases, and even if this problem could in principle be mitigated by hierarchical variants, large-scale applications of whatever kind of fuzzy system are difficult to find.  One of the key advantages of the traditional approach, namely, interpretability and model transparency, tends to become questionable for the data-driven approach.  Moreover, a proper handling of fuzziness in knowledge representation is no longer needed, since data is normally precise, and the alleged fuzziness ``learned'' from the data is  at best suspicious.

Needless to say, our criticism of the traditional rule-based paradigm should not be misunderstood as denying the usefulness of fuzzy logic in machine learning in general. For example, the  model architecture of pattern trees, which were already mentioned at the end of Section 3.1, is an interesting alternative with many appealing properties---the use of generalized (fuzzy) logical and averaging operators is one of the key features of this approach. For a broader discussion of how fuzzy logic can contribute to machine learning, we refer to \cite{mpub223}.


\end{document}